%% file: root.tex
\documentclass[letterpaper, 10 pt, conference]{ieeeconf} 

\IEEEoverridecommandlockouts                        

\usepackage{graphics}
\usepackage{amsmath} 
\usepackage{amssymb}
\usepackage{cite}
\usepackage{tikz}
\usepackage{amsmath}
\usepackage{tabularx}
\usepackage{array}
\usepackage{booktabs}
\usepackage{bm}
\usepackage{algorithm}
\usepackage{algorithmic}
\usepackage{bbm}
\usepackage{adjustbox}
\usepackage{nicematrix}
\usepackage{caption}

\usepackage{tablefootnote}

\usepackage{hyperref}

\definecolor{green}{RGB}{11,155,13}

\DeclareMathOperator*{\argmin}{argmin}

\title{\LARGE \bf
Terrain-Attentive Learning for Efficient 6-DoF Kinodynamic Modeling on Vertically Challenging Terrain
}

\author{Aniket Datar, Chenhui Pan, Mohammad Nazeri, Anuj Pokhrel, and Xuesu Xiao
\thanks{All authors are with the Department of Computer Science, George Mason University {\tt\scriptsize \{adatar, cpan7, mnazerir, apokhre, xiao\}@gmu.edu}}
}

\begin{document}
\maketitle
\thispagestyle{empty}
\pagestyle{empty}

\input{content/abstract.tex}

\input{content/intro.tex}

\input{content/related.tex}
\input{content/approach.tex}

\input{content/experiment.tex}

\input{content/conclusion.tex}

\bibliographystyle{IEEEtran}
\bibliography{IEEEabrv,references}
\end{document}

%% file: content/abstract.tex
\begin{abstract}
Wheeled robots have recently demonstrated superior mechanical capability to traverse vertically challenging terrain (e.g., extremely rugged boulders comparable in size to the vehicles themselves). 
Negotiating such terrain introduces significant variations of vehicle pose in all six Degrees-of-Freedom (DoFs), leading to imbalanced contact forces, varying momentum, and chassis deformation due to non-rigid tires and suspensions. 
To autonomously navigate on vertically challenging terrain, all these factors need to be efficiently reasoned within limited onboard computation and strict real-time constraints. 
In this paper, we propose a 6-DoF kinodynamics learning approach that is attentive only to the specific underlying terrain critical to the current vehicle-terrain interaction, so that it can be efficiently queried in real-time motion planners onboard small robots. 
Physical experiment results show our Terrain-Attentive Learning (\textsc{tal}) demonstrates on average 51.1\% reduction in model prediction error among all 6 DoFs compared to a state-of-the-art model for vertically challenging terrain. 
\end{abstract}

%% file: content/intro.tex
\section{Introduction}

\label{sec::introduction}

Despite their wide availability, wheeled mobile robots are usually limited in terms of mobility, mostly moving in 2D flat environments. After dividing their planar workspaces into free spaces and obstacles, those robots are assumed to be rigid bodies and efficiently find collision-free paths to move from one point to another, using extremely simplified kinodynamic models, e.g., Ackermann-steering or differential-drive. When facing \emph{vertically challenging} terrain, e.g., spaces filled with large obstacles like boulders or tree trunks where a collision-free 2D path does not exist, roboticists have mostly sought help from more sophisticated mechanical design, such as legged, leg-wheeled, and articulated tracked vehicles~\cite{kutzer2010design, bu2022development} or adding active suspension systems~\cite{cordes2014active, jiang2019lateral}. 

Recent advances in wheeled mobility have shown that even conventional wheeled vehicles without sophisticated hardware modification have unrealized mobility potential on vertically challenging terrain~\cite{datar2023toward}. With a set of minimal hardware requirements, e.g., all-wheel drive, independent suspensions, and differential lock, those simple vehicles can also, at least with human teleoperation, venture into environments which would normally be deemed as non-traversable obstacles by state-of-the-art autonomous navigation systems. 

In order to achieve such unrealized mobility potential in an autonomous manner, wheeled robots need to reason about the complex vehicle-terrain interaction, including imbalanced contact forces, varying
momentum, and chassis deformation due to non-rigid tires and suspensions. All these factors are tightly dependent on the underlying terrain. In state-of-the-art motion planners, e.g., sampling-based or optimization-based, such vehicle-terrain interaction needs to be modeled and computed for a large number of future terrain patches beneath candidate vehicle poses. For highly articulated systems, efficient decomposition is possible to break down the modeling of the vehicle chassis and actuators (e.g., legs and active suspensions) so that the chassis trajectory can be computed separately in parallel and the low-level actuation solved using fast control and optimization techniques~\cite{medeiros2020trajectory}. Unfortunately, for under-actuated convectional wheeled robots, the whole system is fully coupled and such decomposition is not possible, requiring sequential, un-parallelizable computation along potential future robot trajectories. 

\begin{figure}
  \centering
  \includegraphics[width=\columnwidth]{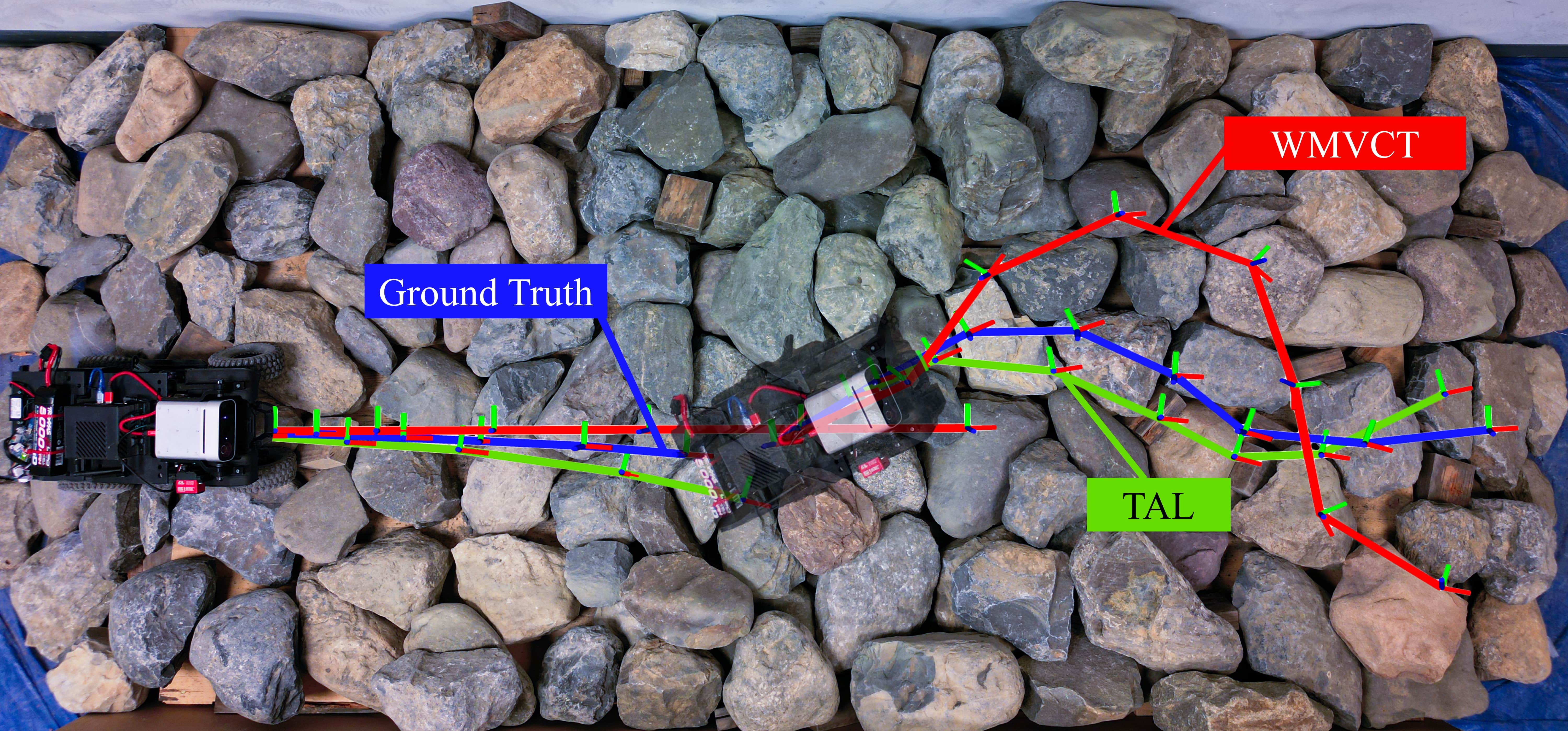}
  \caption{Two Sets of 6-DoF Kinodynamic Trajectory Predictions by \textcolor{green}{\textsc{tal}} and \textcolor{red}{\textsc{wmvct}}~\cite{datar2023learning} Compared to \textcolor{blue}{Ground Truth}.}
  \label{fig::trajectories}
\end{figure}

To this end, we present Terrain-Attentive Learning (\textsc{tal}), a 6-DoF kinodynamics learning approach that is attentive (only) to the specific underlying terrain critical to the current vehicle-terrain interaction, so that it can be efficiently queried in real-time motion planners onboard small robots. \textsc{tal} is combined with a state-of-the-art sampling-based motion planner and allows to sequentially rollout future trajectories in an efficient manner for downstream cost-based kinodynamic planning. Using \textsc{tal}, we demonstrate on average 51.1\% reduction in model prediction error among all 6 DoFs compared to another state-of-the-art kinodynamics modeling approach for vertically challenging terrain~\cite{datar2023learning} (Fig.~\ref{fig::trajectories}).

%% file: content/related.tex
\section{Related Work}
\label{sec::related_work}
This section discusses related work in terms of wheeled robot kinodynamic modeling, off-road navigation, and learning-based mobility. 

\subsection{Wheeled Robot Kinodynamic Modeling}
Wheeled mobile robots have found a variety of real-world applications, e.g., in autonomous delivery~\cite{starship}, warehouse logistics~\cite{amazonrobotics}, scientific exploration~\cite{wettergreen2010science}, and search and rescue~\cite{murphy2014disaster}. 
Thanks to their simplicity and efficiency, differential-drive mechanism~\cite{malu2014kinematics}, Ackermann steering~\cite{atreya2022high}, and omnidirectional wheels~\cite{lu2022design} can efficiently move robots through their planar workspaces~\cite{fox1997dynamic, quinlan1993elastic}, avoid 2D obstacles~\cite{perille2020benchmarking, nair2022dynabarn, xiao2022autonomous}, and reach their goals~\cite{atreya2022high}. 

A simple wheeled robot kinodynamic model is the differential-drive model, i.e., the robot turns from the difference in rotation speed of the left and right wheel(s). Other types of common kinodynamic models for wheeled robots include unicycle, bicycle, and Ackermann-steering model~\cite{lavalle2006planning}, which turn by changing the orientation of the (front) wheel(s). Realizing such extremely simplified models may not be able to account for the imperfectness in the real world, researchers have also developed models with higher physics fidelity, e.g., friction and slip models~\cite{rogers2012aiding, rabiee2019friction}. The benefit of these simple models is that they can be queried in an extremely efficient manner, allowing thousands of potential future model rollouts to be evaluated for downstream kinodynamic planning. 

Most existing wheeled robot kinodynamic models, despite their differences in fidelities, still assume the robot moves in a 2D space and its motion is constrained in $\mathbb{SE}(2)$. However, when facing off-road environments, especially vertically challenging terrain, such an assumption no longer holds and the workspace has to be extended to $\mathbb{SE}(3)$~\cite{datar2023toward, datar2023learning}. Modeling in $\mathbb{SE}(3)$ faces challenges in terms of both accuracy and efficiency: the significant variations of 6-DoF vehicle pose caused by the variety of underlying terrain needs to be precisely modeled, while such a model also needs to be queried efficiently in real-time motion planners. Our \textsc{tal} approach aims at tackling both challenges simultaneously in a data-driven manner using representation learning. 

\subsection{Off-Road Navigation}
A large percentage of off-road navigation research has focused on the perception side since the DARPA Urban Challenge~\cite{seetharaman2006unmanned} and LAGR Program~\cite{jackel2006darpa}. Extending from the simple differentiation of obstacles and free spaces, off-road perception systems need to consider semantic information~\cite{meng2023terrainnet, karnan2023sterling, sikand2022visual, dixit2023step, viswanath2021offseg, maturana2018real, shaban2022semantic}, such as gravel, grass, bushes, pebbles, and rocks, and then devise cost functions based on the semantic understanding for subsequent path and motion planning. 

Recent research efforts have gradually moved towards the mobility side. Inverse~\cite{xiao2021learning, karnan2022vi} and forward~\cite{atreya2022high, maheshwari2023piaug} kinodynamic models have been created from real-world vehicle-terrain interactions~\cite{sivaprakasam2021improving} to enable high-speed off-road navigation. End-to-end learned mobility ~\cite{pan2020imitation} has eliminated the boundary between perception and mobility systems so the whole navigation system can be learned in a data-driven manner. Most existing off-road navigation work still assume the vehicles are moving in a 2D plane, while deliberately choosing which part of the 2D plane to drive on or modeling how different terrain would affect the 3-DoF vehicle motion. 

When facing vertical protrusions from the ground, e.g., large boulders or fallen tree trunks, most existing off-road navigation systems still treat them as non-traversable obstacles, e.g., with a large cost assigned to the corresponding semantic class. In this work, we aim to allow vehicles to efficiently reason about the consequences of interacting with such vertically challenging terrain and autonomously plan feasible motions to traverse through.

\subsection{Learning-Based Mobility}
Recent advancement in machine learning has been utilized for robot mobility~\cite{xiao2022motion} using imitation~\cite{pfeiffer2017perception, xiao2020appld, karnan2022voila} or reinforcement learning~\cite{xu2023benchmarking, xu2021machine, faust2018prm}. Learning enhances robot adaptivity~\cite{xiao2022learning, xiao2022appl, xiao2020appld, wang2021appli, wang2021apple, xu2021applr, liu2021lifelong} and agility~\cite{xiao2021toward, xiao2021agile, wang2021agile}, increases movement speed~\cite{pan2020imitation, xiao2021learning, sivaprakasam2021improving, karnan2022vi, atreya2022high, pokhrel2024cahsor}, enables visual-only navigation~\cite{karnan2022voila, kahn2020badgr, shah2023vint, stachowicz2023fastrlap}, and creates socially compliant mobile robots~\cite{mirsky2021conflict, mavrogiannis2023core, francis2023principles, karnan2022socially, chen2017socially, xiao2022learning, nguyen2023toward, park2023learning, hart2020using}. 

While having the potential to learn from data, learning-based mobility also faces challenges from being data-hungry and computation-intensive, especially onboard a mobile robot. \textsc{tal} aims at alleviating the need of large-scale real-world datasets from constraining the learning process only to a forward kinodynamic model, which will be combined in a Model Predictive Control (MPC)~\cite{williams2016aggressive} setup. \textsc{tal} also utilizes representation learning~\cite{payandeh} so that the learned kinodynamic model can efficiently attend to only the specific underlying terrain critical to the current vehicle-terrain interaction, without extensive computation required to pre-process input data. 

%% file: content/approach.tex
\section{Approach}
\label{sec::approach}

\begin{figure*}
  \centering
  \includegraphics[width=2\columnwidth]{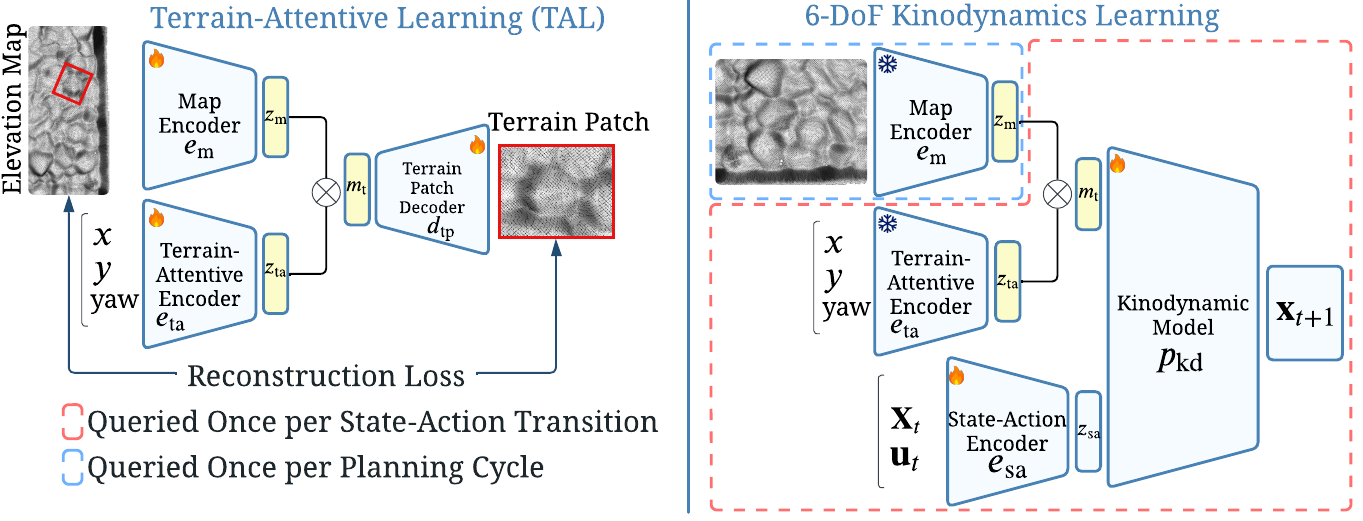}
  \caption{Terrain-Attentive Learning (\textsc{tal}, Left) and 6-DoF Kinodynamics Learning (Right) Architecture: Flame and temperature denote training and frozen parameters respectively. }
  \label{fig::tal}
\end{figure*}

We first formulate the problem of forward kinodynamic modeling for wheeled mobility on vertically challenging terrain. We then present how this problem is approached in a data-driven manner to avoid the need of analytical vehicle-terrain interaction models. Finally, we introduce our \textsc{tal} method which allows the learned 6-DoF kinodynamic model to efficiently attend to the specific underlying terrain so that it can quickly predict the next vehicle state in a MPC setup for downstream kinodynamic planning. 

\subsection{Problem Formulation}
While most traditional 2D navigation problems are defined in a 2D state space, i.e., $X \subset \mathbb{SE}(2)$, our vertically challenging terrain requires the state space to be extended to $X \subset \mathbb{SE}(3)$. Traditional motion planners only move robots in free space and avoid obstacles, as divisions of the whole state space: $X=X_\textrm{free} \cup X_\textrm{obs}$. In contrast, our wheeled robot needs to decide which obstacles should be avoided (as making contact with them will cause immobilization or damage, e.g., hitting a wall), while which ones it can drive on top of (use them as support underneath the chassis), considering a collision-free 2D path may not always exist in vertically challenging environments. 

We adopt a discrete vehicle forward kinodynamic model in the form of 
\begin{equation}
    \mathbf{x}_{t+1} = f_\theta(\mathbf{x}_t, \mathbf{u}_t, \mathbf{m}_t), 
    \label{eqn::model}
\end{equation}
where $\mathbf{x}_t \in X$, $\mathbf{u}_t\in U$, and $\mathbf{m}_t\in M$ denote the vehicle state, control input, and environment state respectively. 
$\mathbf{x}_t$ includes the translations along the $\mathbf{x}$, $\mathbf{y}$, and $\mathbf{z}$ axis ($x$, $y$, and $z$) and the rotations around them (roll, pitch, and yaw) in a coordinate system, as well as their velocity components when necessary. For control input, $\mathbf{u}_t = (v_t, \omega_t) \in U \subset \mathbb{R}^2$, where $v_t$ and $\omega_t$ are the linear and angular velocity or throttle and steering command. 
The environment state $\mathbf{m}_t$ includes all necessary information in the environment to determine the next vehicle state $\mathbf{x}_{t+1}$, given $\mathbf{x}_t$ and $\mathbf{u}_t$. Such information can include terrain geometry and semantics. In this work, we use a 2.5D terrain elevation map to construct $\mathbf{m}_t$ underneath the current vehicle state $\mathbf{x}_t$ to represent terrain topology and leave semantics (e.g., slipperiness, deformability, and elasticity) as future work. 
The motion planning problem is to find a control function $u: \{t\}_{t=0}^{T-1} \rightarrow U$ that produces an optimal path $\mathbf{x}_t \in X_\textrm{free}, \forall t\in \{t\}_{t=0}^T$ from an initial state $\mathbf{x}_0=\mathbf{x}_\textrm{init}$ to a goal region $X_\textrm{goal}\subset X$, i.e., $\mathbf{x}_T\in X_\textrm{goal}$. The path needs to observe the system dynamics $f_\theta(\cdot, \cdot, \cdot)$, parameterized by $\theta$, and minimize a given cost function $c(x)$, which maps from a state trajectory $x: \{t\}_{t=0}^T \rightarrow X$ to a positive real number. 

\subsection{Data-Driven Kinodynamics}
Most existing 2D vehicle kinodynamic models only condition next state $\mathbf{x}_{t+1}$ on current state $\mathbf{x}_t$ and input $\mathbf{u}_t$ and are significantly simplified using, e.g., differential-drive, unicycle, bicycle, or Ackermann-steering mechanisms. However, the inclusion of $\mathbf{m}_t$ when moving in off-road environments, especially on vertically challenging terrain, substantially complicates the model. For example, driving the vehicle toward a wall or an extremely large slope will get the vehicle stuck; driving quickly on undulating terrain may cause the vehicle to be airborne; Driving on extremely slanted terrain may compromise vehicle stability and lead to rollover. How different terrain characteristics may affect the vehicle-terrain interaction is very difficult to analytically model. 

To avoid the difficulty in analytically modeling $f_\theta$, we adopt a data-driven approach. We assume a training dataset of size $N$ is available: $\mathcal{D} = \{ \langle \mathbf{x}_t, \mathbf{x}_{t+1}, \mathbf{m}_t, \mathbf{u}_t \rangle_{t=1}^N$\}. $\theta$ can then be learned by minimizing a supervised loss function: 
\begin{equation}
    \theta^* = \argmin_{\theta} \sum_{(\mathbf{x}_{t}, \mathbf{x}_{t+1}, \mathbf{m}_t, \mathbf{u}_t) \in \mathcal{D}} \lVert f_\theta(\mathbf{x}_{t}, \mathbf{u}_t, \mathbf{m}_t) - \mathbf{x}_{t+1}\rVert,
    \label{eqn::bc}
\end{equation}
The learned vehicle-terrain forward kinodynamic model $f_\theta(\cdot, \cdot, \cdot)$, e.g., instantiated as a deep neural network, can be used to rollout future trajectories for minimal-cost planning. 

\begin{figure*}
  \centering
  \includegraphics[width=2.05\columnwidth]{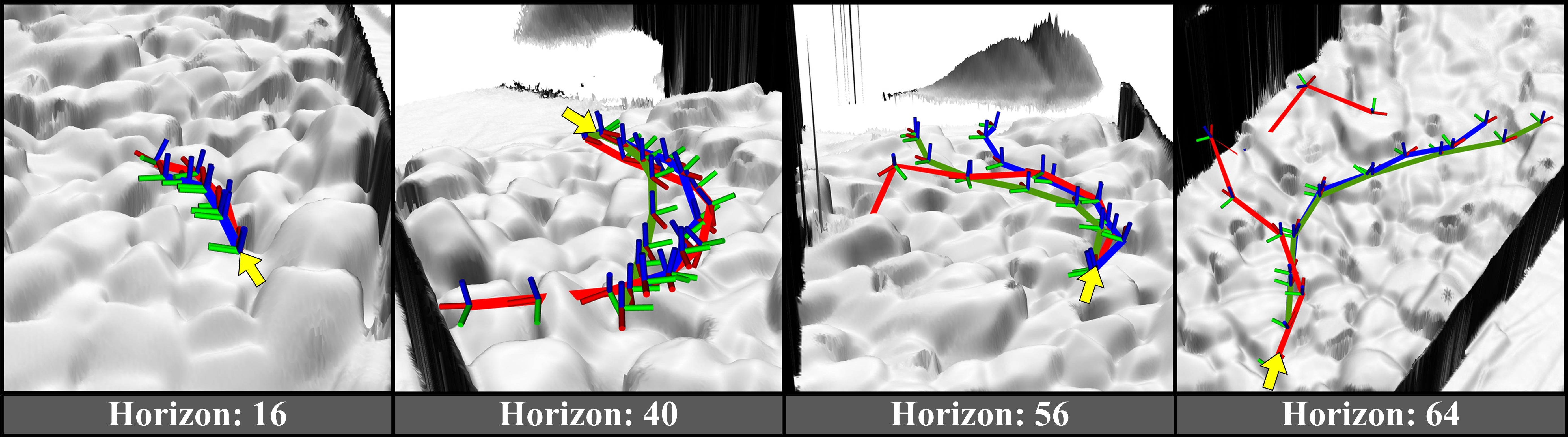}
  \caption{6-DoF Vehicle Trajectories of \textcolor{green}{\textsc{tal}}, \textcolor{red}{\textsc{wmvct}}, and \textcolor{blue}{Ground Truth} with Increasing Horizon: \textcolor{green}{\textsc{tal}} closely matches \textcolor{blue}{Ground Truth} even with a long horizon, while \textcolor{red}{\textsc{wmvct}} significantly diverges. }
  \label{fig::horizon_trajectories}
\end{figure*}

\subsection{Terrain-Attentive Learning (\textsc{tal})}
In addition to the difficulty in precisely deriving analytical models for $f_\theta$, another difficulty brought by the inclusion of $\mathbf{m}_t$ is the increased computation cost and reduced efficiency during model query. 
The state-action transitions of simplified 2D kinodynamic models, when depending only on the current state and input, can therefore be very quickly computed. They can even be pre-computed and saved in advance, e.g., as state lattices~\cite{pivtoraiko2009differentially} or pre-processed maps~\cite{cai2022risk}. Conversely, even given the same current state $\mathbf{x}_t$ and input $\mathbf{u}_t$, different $\mathbf{m}_t$ as input will produce a variety of next state $\mathbf{x}_{t+1}$, which will further affect the transition into $\mathbf{x}_{t+2}$, and so on. 
In a MPC setup, such a sequential dependence of the next state-action transition on the current one precludes the possibility of processing the sequence of $\{\mathbf{m}_t\}_{t=0}^{H-1}$ for one single trajectory (with $H$ as the planning horizon) in parallel and therefore incurs extensive computation overhead during sequential rollouts, especially when a large amount of potential state-action transitions must be computed for iterative, sampling-based motion planners~\cite{williams2016aggressive}.  
Furthermore, how to efficiently extract $\mathbf{m}_t$ from raw perception within limited onboard computation is also a challenging task. 

Therefore, \textsc{tal} utilizes self-supervised representation learning to efficiently process robot perception into $\mathbf{m}_t$ (Fig.~\ref{fig::tal} left) and query the learned model $f_\theta$ (Fig.~\ref{fig::tal} right) in order to rollout and evaluate future candidate trajectories. 
Within a MPC planning cycle, the kinodynamic model needs to quickly retrieve relevant environment state from the space of all possible environment states, i.e., $\mathbf{m}_t\in M$. 
In our wheeled mobility on vertically challenging terrain problem, $M$ is the terrain information space of all possible terrain patches that can be extracted from an elevation map built by an online mapping system~\cite{mikielevation2022}. 
Given a full 2.5D elevation map $E$ of the vertically challenging terrain, the terrain patch underneath the robot state $\mathbf{x}_t$ can be extracted using the vehicle $x_t$, $y_t$, and $\textrm{yaw}_t$, along with the constant vehicle footprint. Notice that such terrain extraction requires translation, cropping, and rotation operations of the original full elevation map and therefore incurs an extensive amount of computation when repeated many times in a sampling-based MPC setting. Furthermore, consuming the terrain patch as kinodynamic model input during every state-action transition is also extremely computationally extensive. 
To use representation learning to alleviate the computation overhead during deployment, we generate a terrain patch dataset using many full elevation maps $\{E_i\}_{i=1}^I$ and terrain patches extracted from each of them based on randomly sampled $\langle x, y, \textrm{yaw} \rangle$ tuples, denoted as $\{E_i, \{p_i^j, \langle x_i^j, y_i^j, \textrm{yaw}_i^j\rangle\}_{j=1}^J\}_{i=1}^I$. As shown in Fig.~\ref{fig::tal} left, a map encoder $e_\textrm{m}$ and a terrain-attentive encoder $e_\textrm{ta}$ embed the full elevation map $E$ and $\langle x, y, \textrm{yaw} \rangle$ into their latent spaces, before being concatenated and decoded using a terrain patch decoder $d_\textrm{tp}$. The map and terrain-attentive encoders and the terrain patch decoder are trained in an end-to-end fashion using self-supervised representation loss:
\begin{equation}
    \mathcal{L}_\textsc{tal} = \sum_{i=1}^I \sum_{j=1}^J \lVert p_i^j - d_\textrm{tp}(e_\textrm{m}(E_i), e_\textrm{ta}(\langle x_i^j, y_i^j, \textrm{yaw}_i^j \rangle))\rVert. 
\end{equation}
The latent embeddings of the full elevation map and  $\langle x, y, \textrm{yaw} \rangle$ contain sufficient information to reconstruct the terrain patch, and therefore can be used as $\mathbf{m}_t$. 

The parameters for the learned map and terrain-attentive encoders, $e_\textrm{m}$ and $e_\textrm{ta}$, are then frozen during downstream 6-DoF kinodynamics learning (Fig.~\ref{fig::tal} right). The optimal kinodynamics parameters $\theta^*$, in the form of a state-action encoder $e_\textrm{sa}$ and kinodynamics predictor $p_\textrm{kd}$, are learned using the kinodynamics loss defined in Eqn.~\eqref{eqn::bc}. During a single deployment planning cycle, the large map encoder will only need to be queried once and produce one elevation map embedding, while the small terrain-attentive encoder, state-action encoder, and kinodynamics predictor will be queried for every state-action transition. The learned kinodynamic model can then be efficiently queried for subsequent sampling-based MPC planning.  

\subsection{Implementations}
\subsubsection{Terrain Attentive Learning} \textsc{tal} leverages a 3-layer Convolutional Neural Network (CNN) as the map encoder ($e_\textrm{m}$) that produces a latent embedding $\mathbf{z}_{\text{m}} \in \mathbb{R}^{160 \times 6 \times 6}$.
In parallel, the terrain-attentive encoder ($e_\textrm{ta}$), a 2-layer Multi-Layer Perceptron (MLP), produces a latent embedding $\mathbf{z}_{\text{ta}} \in \mathbb{R}^{160 \times 6 \times 6}$, the same size as $\mathbf{z}_{\text{m}}$. The second embedding $\mathbf{z}_{\text{ta}}$ serves as attention weights, which are subsequently multiplied with $\mathbf{z}_{\text{m}}$ and passed through one linear layer producing a latent embedding $\mathbf{m}_t \in \mathbb{R}^{64 \times 6 \times 6}$ as the final terrain representation.
The terrain patch decoder $d_\textrm{tp}$ is a 4-layer Convolutional Transpose Network to reconstruct the patch corresponding to the robot footprint with a $0.24\text{m}^2$ area in the real world.
We use Mean Squared Error as the loss function to guide the reconstruction process. 

\subsubsection{6-DoF Kinodynamics Learning} The kinodynamics learning consists of the pre-trained \textsc{tal} model with the addition of the state-action encoder $e_\textrm{sa}$ and the kinodynamics predictor $p_\textrm{kd}$.
The state-action encoder $e_\textrm{sa}$ incorporates two MLPs each with two layers to encode state ($\mathbf{x}_t$) into $\mathbf{z}_\text{s} \in \mathbb{R}^{16}$ and action ($\mathbf{u}_t$) into $\mathbf{z}_\text{a} \in \mathbb{R}^{16}$.
Then we concatenate $\mathbf{z}_\text{s}$ and $\mathbf{z}_\text{a}$ into $\mathbf{z}_\text{sa}$.
This $\mathbf{z}_{\text{sa}}$ is then further concatenated with the terrain representation, $\mathbf{m}_t$, obtained from the \textsc{tal} model. 

The concatenated vector, consisting of $\mathbf{z}_{\text{sa}}$ and $\mathbf{m}_t$, is subsequently fed into the kinodynamics predictor $p_\textrm{kd}$, a 2-layer MLP, to predict the next state $\mathbf{x}_{t+1}$. 
During this stage, the weights of $e_\textrm{m}$ and $e_\textrm{ta}$ are frozen, and only the weights of $e_\textrm{sa}$ and $p_\textrm{kd}$, i.e., $f_{\theta}$, are updated through training.

%% file: content/experiment.tex
\section{Experiments}
\label{sec::experiments}

\begin{figure*}
  \centering
  \includegraphics[height=0.53\columnwidth]{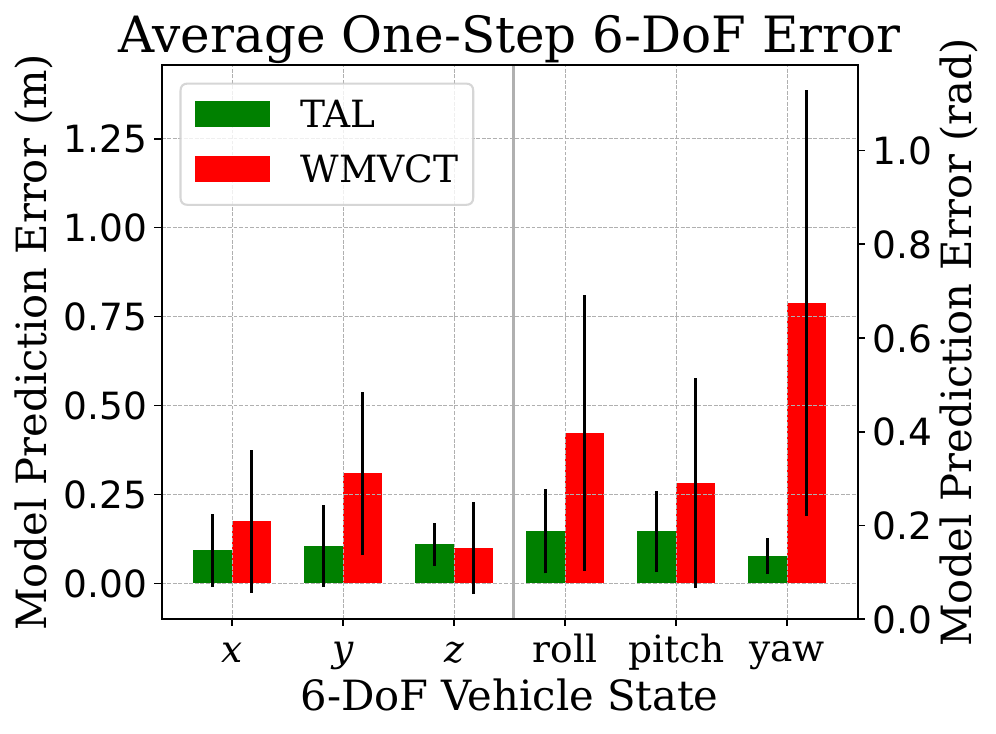}
  \includegraphics[height=0.53\columnwidth]{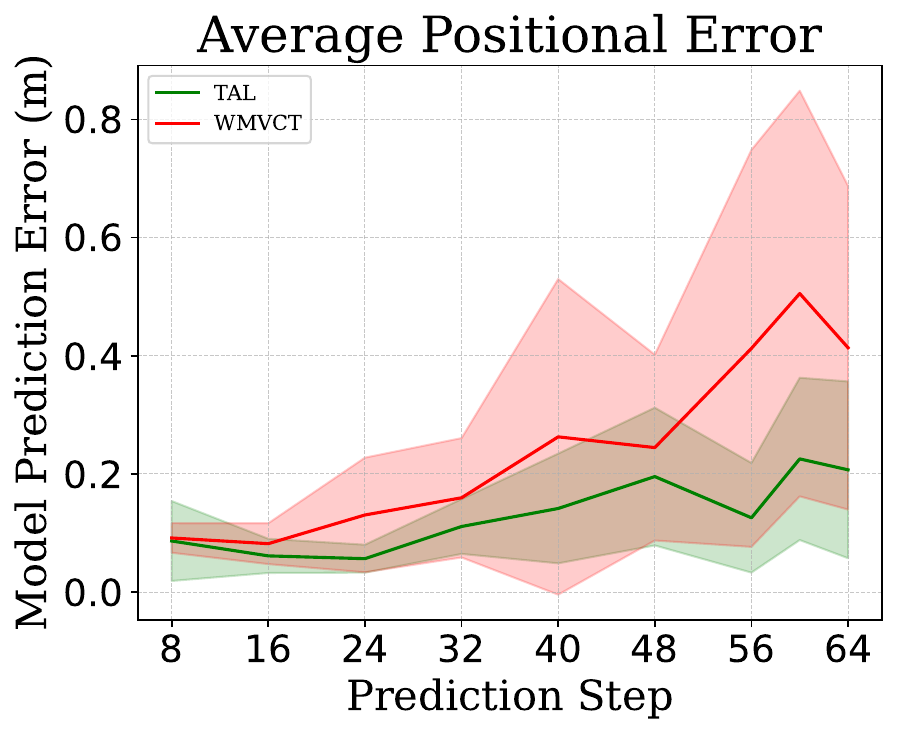}
  \includegraphics[height=0.53\columnwidth]{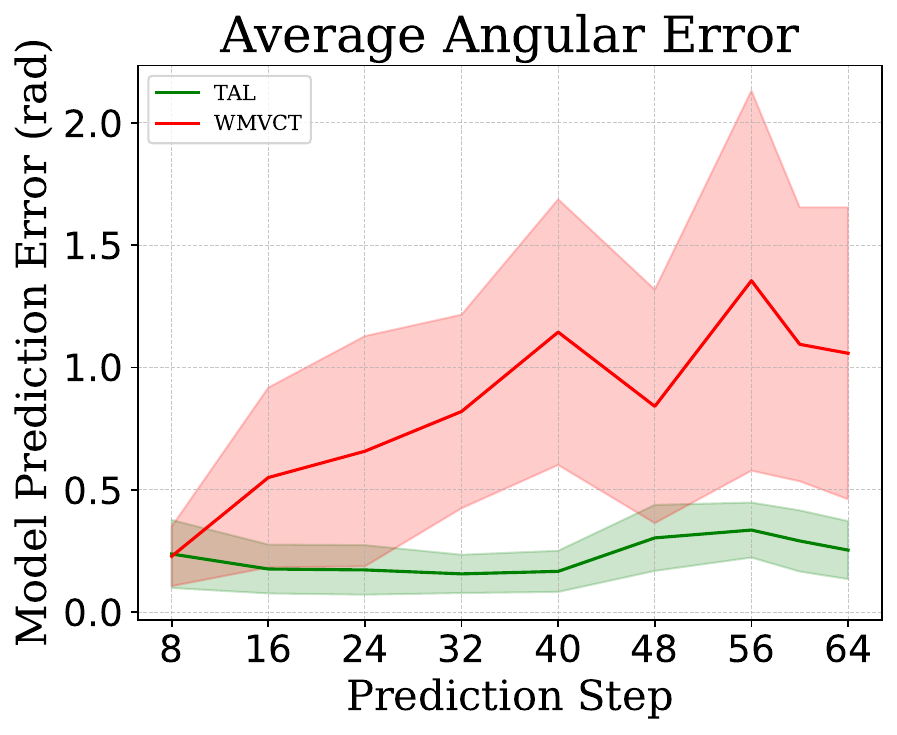} 
  \caption{Model Prediction Error of \textcolor{green}{\textsc{tal}} and \textcolor{red}{\textsc{wmvct}}: Average One-Step 6-DoF Positional and Angular Error (Left);  Prediction Error vs. Prediction Step (Middle and Right). \textcolor{green}{\textsc{tal}} achieves lower prediction error and variance than \textcolor{red}{\textsc{wmvct}} in all cases. }
  \label{fig::error}
\end{figure*}

We conduct experiments to verify that our \textsc{tal} model is able to produce accurate future vehicle state prediction based on the current state $\mathbf{x}_t$, current action $\mathbf{u}_t$, and underlying terrain $\mathbf{m}_t$. We compare the prediction from \textsc{tal} against another state-of-the-art 6-DoF vehicle kinodynamic model for vertically challenging terrain used in the \textsc{wmvct} planner~\cite{datar2023learning}. We also deploy the \textsc{tal} model in a sampling-based planner and show it can be used to genreate feasible motion plans to navigate through vertically challenging terrain.

\subsection{Robot, Testbed, and Data}
We implement \textsc{tal} on an open-source, 1/10th-scale, unmanned ground vehicle, the Verti-4-Wheeler (V4W) platform~\cite{datar2023toward}. 
The robot is equipped with a low-high gear switch and lockable front and rear differentials enhancing its mobility on vertically challenging terrain. 
For simplicity, in our datasets and experiments, we only use low-gear and always lock both differentials and leave the investigation of the effect of low/high gear and locked/released differentials on kinodynamics to future work. 
For perception, we use the onboard Microsoft Azure Kinect RGB-D camera and perform Visual Inertia Odometry (VIO)~\cite{chen2023direct}. 
We use an open-source tool to build real-time elevation map~\cite{miki2022elevation} based on the depth input. 
For computation, an NVIDIA Jetson Orin NX computer is available onboard. 

We construct a 3.1m$\times$1.3m rock testbed with a maximum height of 0.6m (Fig.~\ref{fig::trajectories}). For comparison, the V4W has a height of $0.2$m, width of $0.249$m, and length of $0.523$m with a $0.312$m wheel base.  
The numerous rocks on this rock testbed can be easily reconfigured in order to facilitate data collection and mobility experiments in a wide variety of configurations. 

\begin{figure}
  \centering
  \includegraphics[width=0.75\columnwidth]{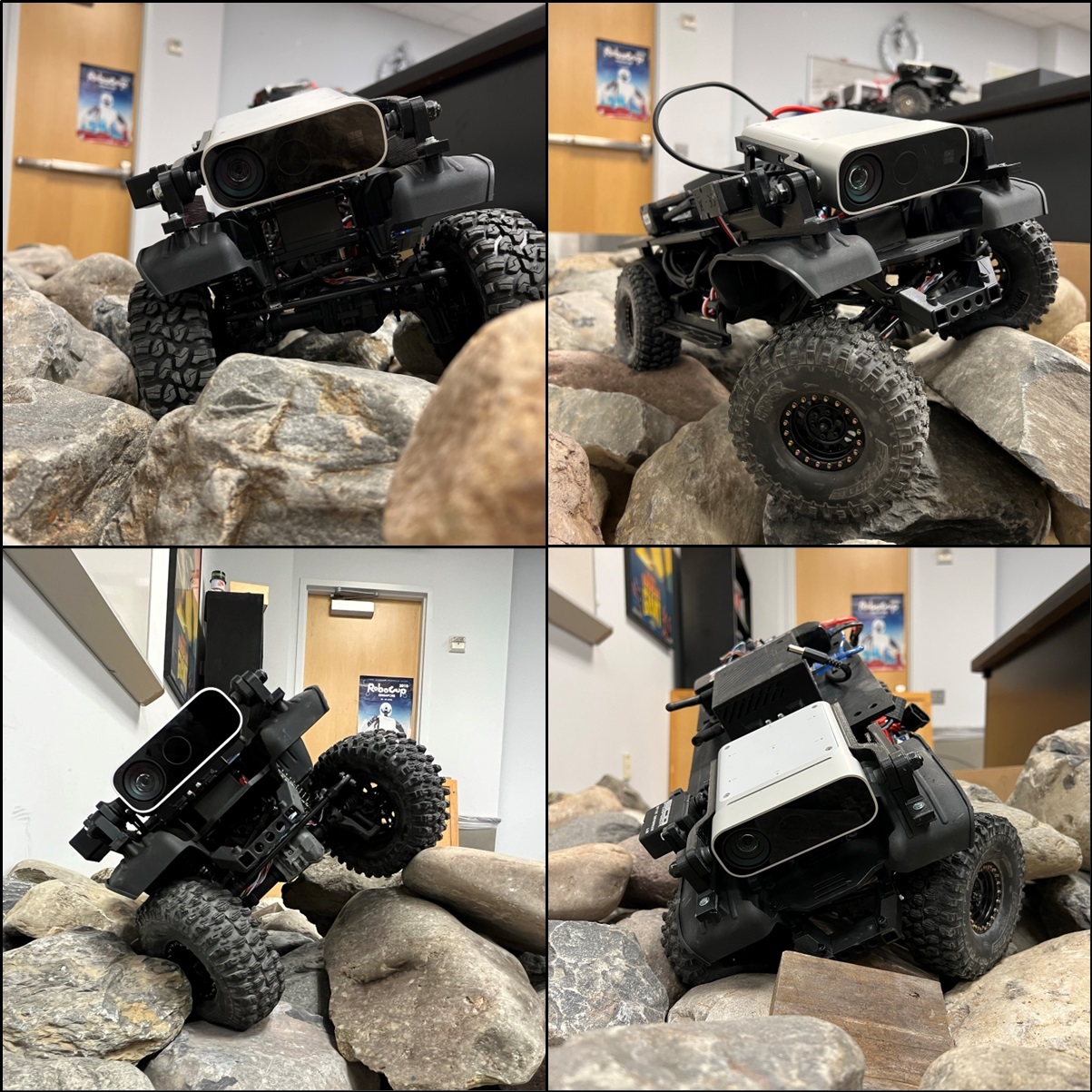}
  \caption{Diverse 6-DoF Vehicle States in the Dataset.}
  \label{fig::dataset}
\end{figure}

We collect 30 minutes of data on the rock testbed and 30 minutes of data on a planar surface. We use a 9:1 ratio to split train and test data and report all results on unseen test data. The dataset contains VIO for vehicle state estimation, elevation maps built from depth images, and teleoperated vehicle controls including throttle and steering commands. A variety of 6-DoF vehicle states are included in the rock testbed data, including vehicle rollover and getting stuck. Example 6-DoF vehicle states are shown in Fig.~\ref{fig::dataset}. 

\subsection{Trajectory Prediction Visualization}
Fig.~\ref{fig::horizon_trajectories} visualizes a set of predicted trajectory examples by \textsc{tal} and \textsc{wmvct} compared against the ground truth at different horizon steps. At horizon 16, all three trajectories are close to each other; At horizon 40, \textsc{wmvct} fails to consider the resistance from a large rock and reaches much farther than \textsc{tal}, whose length is similar to the ground truth; At horizon 56, \textsc{tal} follows the ground truth direction in general, while \textsc{wmvct} deviates to the left; At horizon 64, significant error is accumulated by \textsc{wmvct}, causing the red trajectory reaching out of the elevation map and then penetrating the rocks, but \textsc{tal} still closely follows the ground truth. 

\subsection{6-DoF Prediction Accuracy}
We compare the accuracy of the \textsc{tal} model in predicting the next 6-DoF vehicle state with the model used in \textsc{wmvct}~\cite{datar2023learning}. For efficiency, the \textsc{wmvct} model decomposes the 6 DoFs into three parts: $x$, $y$, and yaw are determined by a simple planar Ackermann-steering model; $z$ is based on the elevation map value at $(x, y)$; roll and pitch are computed using a neural network which takes as input a terrain patch located at $(x, y)$ and aligned with yaw. 
Fig.~\ref{fig::error} left shows the average error with standard deviation in predicting the 6-DoF vehicle state.
Except the negligible difference in $z$-position of the robot, \textsc{tal} outperforms \textsc{wmvct} for all other DoFs by a wide margin, with significantly smaller variance. 
Averaged among all DoFs, \textsc{tal} achieves 51.1\% reduction in modeling error and 62.5\% reduction in error standard deviation. 
Fig.~\ref{fig::error} middle and right shows the 6-DoF prediction error of the models with respect to different prediction steps. 
With increasing steps, error significantly accumulates for \textsc{wmvct} and the increasing variance indicates higher uncertainty, while \textsc{tal} can predict the positions as well as the angles with more accuracy and smaller variance.

\subsection{On-Robot Deployment}
We deploy \textsc{tal} with the Model Predictive Path Integral (\textsc{mppi}) planner~\cite{williams2017model} on V4W. 
The \textsc{mppi} planner operates by rolling out 400 candidate trajectories at each time step, extending its planning horizon to 20 steps into the future.
For sampling diverse control sequences, the \textsc{mppi} planner uses a normal distribution centered around the actual control sequence executed in the last time step by the robot.
This set of candidate control sequences, along with the elevation map, is then fed into the \textsc{tal} model. 
For each time step within each trajectory, \textsc{tal} predicts the resulting 6-DoF state of the robot based on the initial or the last predicted state. 
These resultant states are then fed to a custom cost function, which takes into account the Euclidean distance to the goal along with the roll and pitch values of the predicted states. 
The cost function penalizes the states with high roll and pitch values, which incentivizes the robot to prioritize trajectories that maintain stable vehicle poses while approaching the goal.
The performance of the \textsc{mppi} in conjunction with \textsc{tal} is assessed on unseen test rock configurations after shuffling the rock testbed. 
To be specific, we manually increase the navigation difficulty by introducing ``tricky corners'' for the robot to avoid in order to maintain low roll and pitch angles. 
We then conduct a series of experiments, running 10 trials each for \textsc{mppi} with \textsc{tal}, the \textsc{wmvct} planner~\cite{datar2023learning}, and Behavior Cloning (BC)~\cite{pomerleau1989alvinn, bojarski2016end, nazeri} using the same training data, as well as two baselines provided by the Verti-Wheelers project, i.e., Rule-Based (\textsc{RB}) and Open-Loop (OL)~\cite{datar2023toward}. To be specific, the \textsc{wmvct} planner uses a fixed set of vehicle state trajectory rollouts, which are not dependent on the vehicle action but a set of pre-determined arcs, and then employs a PID controller to track such state trajectories. 
The goal is consistently set across the rock testbed for all trials. 

\begin{table}
\centering
\caption{Comparison of Success Rate and Average Time.}
\label{tab::physical}
\begin{NiceTabular}[columns-width = 0.25cm,rules/width=1pt]{@{}>{\bfseries}cccW{c}{35pt}W{c}{30pt}W{c}{40pt}@{}}
\toprule
 & OL & RB & BC & \textsc{wmvct} & \textsc{tal}\\
\midrule
Success Rate & 0/10 & 0/10 & 7/10 & \textbf{10/10} & \textbf{10/10}\\
Average Time & - & - & 12.28$\pm$2.69 & 16.76$\pm$1.44 & \textbf{16.53$\pm$1.08} \\

\bottomrule
\end{NiceTabular}
\end{table}

Table~\ref{tab::physical} presents the success rate and average traversal time (for successful trials) of all five methods. The ``tricky corners'' cause trouble for OL and RB every time and the V4W either gets stuck or rolls over, achieving zero successful trials. BC fails three trials due to the same reasons. \textsc{wmvct} with the decomposed 6-DoF model performs similarly as \textsc{mppi} with \textsc{tal}. The significantly higher accuracy does not directly translate to much better navigation performance. 

\subsection{Discussions}
In our experiments, the \textsc{tal} model achieves significantly better prediction accuracy compared to the \textsc{wmvct} model in all six DoFs and does not accumulate extensive error during long-horizon prediction. However, such a superior model accuracy does not translate to higher success rate when being used in the \textsc{mppi} planner. 
We posit that the reason for a missing direct correlation between significantly higher model accuracy and better navigation performance is the \textsc{mppi} planner's extensive computation demands. While the \textsc{wmvct} planner is able to quickly update the plan using an extremely efficient but inaccurate model, the \textsc{mppi} planner takes longer to converge when using the high-accuracy \textsc{tal} model. This increased computational cost leads to a reduction in the planning frequency, which further hinders the robot's ability to react and avoid risky obstacles in time. Such an observation motivates future investigation into the tradeoff between high model fidelity and planning frequency.

%% file: content/conclusion.tex
\section{Conclusions}
This work introduces Terrain Attentive Learning (\textsc{tal}) for 6-DoF kinodynamics learning, focusing on extracting important features that influence robot-terrain interaction.
Specifically, we pre-train neural networks to use robot poses as attention weights. 
These attention weights guide the extraction of important underlying features from the elevation map, utilizing patch reconstruction as a form of self-supervision.  
With the pre-trained networks, \textsc{tal} predicts the next vehicle state based on the current pose, control input, and elevation map. 
This approach enables efficient deployment in real-time planners for small robots with limited computational resources.
We quantitatively and qualitatively show that \textsc{tal} can accurately predict the next robot state, which helps to plan feasible, stable, and efficient paths through vertically challenging terrain in a sampling-based motion planner. 
\label{sec::conclusions}